%% file: main.tex
\documentclass[nosubfloats]{hld2023} %

\usepackage[dvipsnames]{xcolor}
\usepackage{tikz}

\usetikzlibrary{calc,patterns,decorations.pathreplacing,shapes.multipart}
\usepackage{float}  

\usepackage{subfig}
\usepackage[justification=justified]{caption}

\usepackage{todonotes}

\input{macros.tex}

\title[Implicit regularisation in stochastic gradient descent]{Implicit regularisation in stochastic gradient descent:\\ from single-objective to two-player games}

\hldauthor{%
\Name{Mihaela Rosca} \Email{mihaelacr@google.com}\\
\addr {Google DeepMind, University College London} 
\AND
\Name{Marc Deisenroth} \Email{m.deisenroth@ucl.ac.uk}\\
\addr {University College London} 
}

\begin{document}

\maketitle

\begin{abstract}%
Recent years have seen many insights on deep learning optimisation being brought forward by finding implicit regularisation effects of commonly used gradient-based optimisers~\citep{igr,igr_sgd,vardi2021implicit,gunasekar2018implicit,schafer2019implicit,neyshabur2014search,blanc2020implicit,keskar2016large,cohen2021gradient,razin2020implicit}. Understanding implicit regularisation can not only shed light on optimisation dynamics, but it can also be used to improve performance and stability across problem domains, from supervised learning to two-player games such as Generative Adversarial Networks~\citep{igr,igr_sgd,geiping2021stochastic,rosca2021discretisation,schafer2019implicit}.
An avenue for finding such implicit regularisation effects has been quantifying the discretisation errors of discrete optimisers via continuous-time flows constructed by backward error analysis (BEA)~\citep{hairer2006geometric}. The current usage of BEA is not without limitations, since not all the vector fields of continuous-time flows obtained using BEA can be written as a gradient, hindering the construction of modified losses revealing implicit regularisers. In this work, we provide a novel approach to use BEA, and show how our approach can be used to construct continuous-time flows with vector fields that can be written as gradients. We then use this to find previously unknown implicit regularisation effects, such as those induced by multiple stochastic gradient descent steps while accounting for the exact data batches used in the updates, and in generally differentiable two-player games.
\end{abstract}

\section{Revisiting Backward error analysis}
\label{sec:bea}

Given a loss $E: \mathbb{R}^D \rightarrow \mathbb{R}$, the gradient descent (GD) update with learning rate $h$, namely ${\vtheta_t = \vtheta_{t-1} - h\nabla_{\vtheta} E(\vtheta)}$, is obtained via Euler discretisation of the gradient flow ${\dot{\vtheta} =   - \nabla_{\vtheta} E}$. After one GD step, the discretisation error $\| \vtheta_t - \vtheta(h; \vtheta_{t-1})\|$ is of order $\mathcal O(h^{2})$, where $\vtheta(h; \vtheta_{t-1})$ is the solution of the flow at time $h$ with $\vtheta(0) = \vtheta_{t-1}$. 
Backward error analysis (BEA)~\citep{hairer2006geometric} provides a technique to quantify this discretisation error, by finding
$f_i: \mathbb{R}^D \rightarrow \mathbb{R}^D$, such that the flow
\begin{align}
    \bm{{\dot{\vtheta}}} =  - \nabla_{\vtheta} E + h f_1(\vtheta) + \cdots + h^n f_n(\vtheta)
\label{eq:general_modified_vector_field}
\end{align} 
follows the GD update with an error $\| \vtheta_t - \vtheta(h; \vtheta_{t-1})\|$ of order $\mathcal O(h^{n+2})$.
\citet{igr} used this technique to find $f_1$, thus finding the flow with an error of $\mathcal{O}(h^3)$ after one GD update:
\begin{align} 
\dot{\vtheta} = -\nabla_{\vtheta}E   -\frac{h}{2} \nabla_{\vtheta}^2 E \nabla_{\vtheta} E  = -\nabla_{\vtheta}\left(E(\vtheta) + \frac h4 \norm{\nabla_{\vtheta} E(\vtheta)}^2\right).
\label{eq:first_igr}
\end{align}
GD can thus be seen as implicitly minimising the modified loss $E(\vtheta) + \frac h4 \norm{\nabla_{\vtheta} E(\vtheta)}^2$.
This showcases an implicit regularisation effect induced by the discretisation error of GD, dependent on learning rate $h$, which biases learning towards paths with low gradient norms. The authors refer to this phenomenon as `implicit gradient regularisation'; we refer to the flow in Eq~\eqref{eq:first_igr} as the IGR flow.

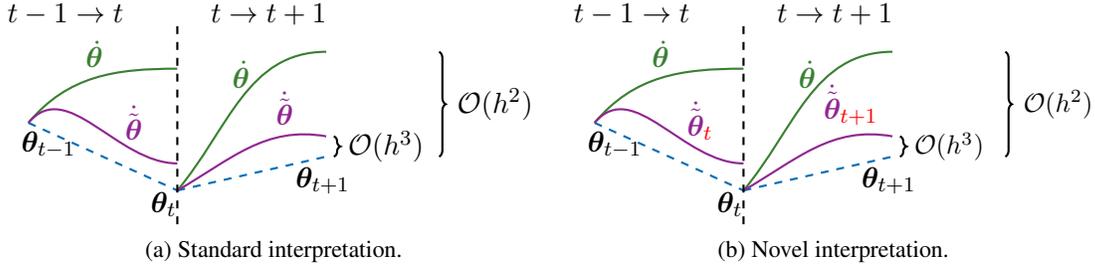
\begin{figure}[t]
\subfloat[Standard interpretation.]{
\begin{tikzpicture}[every text node part/.style={align=center,inner sep=0,outer sep=0}, scale=0.9][overlay]
\coordinate (theta_t_minus_1) at (0,0);
\coordinate (theta_t) at (2.2,-1);
\coordinate (theta_t_plus_1) at (4.4, -0.5);

\node(draw) at ($(theta_t_minus_1) + (+0.3,-0.32)$) {$\vtheta_{t-1}$};
\node(draw) at ($(theta_t) + (-0.2,-0.2)$) {$\vtheta_{t}$};
\node(draw) at ($(theta_t_plus_1) + (-0.05,-0.35)$) {$\vtheta_{t+1}$};

\coordinate (first_time_transition) at ($(theta_t_minus_1) + (+0.55,1.6)$);
\coordinate (second_time_transition) at ($(first_time_transition) + (3,0)$);

\node(draw) at (first_time_transition) {$t -1 \rightarrow t $};
\node(draw) at (second_time_transition) {$t \rightarrow t + 1$};

\coordinate (cont_theta_t) at (2.2,0.8);
\coordinate (cont_theta_t_plus_1) at (4.4, 1.05);

\coordinate (mod_cont_theta_t) at (2.2, -0.6);
\coordinate (mod_cont_theta_t_plus_1) at (4.4, -0.2);

\draw [NavyBlue,thick,dashed] (theta_t_minus_1) -- (theta_t);
\draw [NavyBlue,thick,dashed] (theta_t) -- (theta_t_plus_1);

\draw [OliveGreen,thick]  (theta_t_minus_1) to[out=50,in=180] node[midway,above] {$\dot{\vtheta}$} (cont_theta_t);
\draw [OliveGreen,thick]  (theta_t) to[out=50,in=180] node[midway,above]{$\dot{\vtheta}$} (cont_theta_t_plus_1);

\draw [Plum,thick]  (theta_t_minus_1) to[out=50,in=180]  node[near end,above] {$\dot{\tilde{\vtheta}}$} (mod_cont_theta_t);
\draw [Plum,thick]  (theta_t) to[out=30,in=170] node[near end,above] {$\dot{\tilde{\vtheta}}$} (mod_cont_theta_t_plus_1);

\draw [black,thick,dashed] ($(theta_t) + (0,-0.5)$) -- ($(cont_theta_t) + (0,0.7)$);

\draw [
    thick,
    decoration={
        brace,
        mirror,
        raise=0.1cm
    },
    decorate
] (theta_t_plus_1) -- (mod_cont_theta_t_plus_1)
node [pos=0.5,anchor=west,xshift=0.15cm] {$\mathcal{O}(h^3)$};

\draw [
    thick,
    decoration={
        brace,
        mirror,
        raise=1.5cm
    },
    decorate
] (theta_t_plus_1) -- (cont_theta_t_plus_1)
node [pos=0.5,anchor=west,xshift=1.6cm] {$\mathcal{O}(h^2)$};
\vspace{-2em}
\end{tikzpicture}
\label{tkz:orig_bea}
}
\subfloat[Novel interpretation.]{
\begin{tikzpicture}[every text node part/.style={align=center,inner sep=0,outer sep=0},scale=0.9][overlay]
\coordinate (theta_t_minus_1) at (0,0);
\coordinate (theta_t) at (2.2,-1);
\coordinate (theta_t_plus_1) at (4.4, -0.5);

\node(draw) at ($(theta_t_minus_1) + (+0.3,-0.32)$) {$\vtheta_{t-1}$};
\node(draw) at ($(theta_t) + (-0.2,-0.2)$) {$\vtheta_{t}$};
\node(draw) at ($(theta_t_plus_1) + (-0.05,-0.35)$) {$\vtheta_{t+1}$};

\coordinate (first_time_transition) at ($(theta_t_minus_1) + (+0.55,1.6)$);
\coordinate (second_time_transition) at ($(first_time_transition) + (3,0)$);

\node(draw) at (first_time_transition) {$t -1 \rightarrow t $};
\node(draw) at (second_time_transition) {$t \rightarrow t + 1$};

\coordinate (cont_theta_t) at (2.2,0.8);
\coordinate (cont_theta_t_plus_1) at (4.4, 1.05);

\coordinate (mod_cont_theta_t) at (2.2, -0.6);
\coordinate (mod_cont_theta_t_plus_1) at (4.4, -0.2);

\draw [NavyBlue,thick,dashed] (theta_t_minus_1) -- (theta_t);
\draw [NavyBlue,thick,dashed] (theta_t) -- (theta_t_plus_1);

\draw [OliveGreen,thick]  (theta_t_minus_1) to[out=50,in=180] node[midway,above] {$\dot{\vtheta}$} (cont_theta_t);
\draw [OliveGreen,thick]  (theta_t) to[out=50,in=180] node[midway,above]{$\dot{\vtheta}$} (cont_theta_t_plus_1);

\draw [Plum,thick]  (theta_t_minus_1) to[out=50,in=180]  node[near end,above] {$\dot{\tilde{\vtheta}}_{\color{red}{t}}$} (mod_cont_theta_t);
\draw [Plum,thick]  (theta_t) to[out=30,in=170] node[near end,above] {$\dot{\tilde{\vtheta}}_{{\color{red}{t + 1}}}$} (mod_cont_theta_t_plus_1);

\draw [black,thick,dashed] ($(theta_t) + (0,-0.5)$) -- ($(cont_theta_t) + (0,0.7)$);

\draw [
    thick,
    decoration={
        brace,
        mirror,
        raise=0.1cm
    },
    decorate
] (theta_t_plus_1) -- (mod_cont_theta_t_plus_1)
node [pos=0.5,anchor=west,xshift=0.15cm] {\small $\mathcal{O}(h^3)$};

\draw [
    thick,
    decoration={
        brace,
        mirror,
        raise=1.5cm
    },
    decorate
] (theta_t_plus_1) -- (cont_theta_t_plus_1)
node [pos=0.5,anchor=west,xshift=1.6cm] {\small $\mathcal{O}(h^2)$};
\vspace{-2em}
\end{tikzpicture}
\label{tkz:novel_bea}
}
   \caption[Visualising the standard approach to backward error analysis alongside an approach which constructs a different flow per gradient descent iteration.]{ Previous uses of BEA \subref{tkz:orig_bea} construct modified flows $\dot{\tilde{\vtheta}}$ to capture the discretisation error of updates obtained by discretising $\dot{\vtheta}$; these flows did not depend on the initial iteration parameters. Here, we take the second approach \subref{tkz:novel_bea}, allowing us to construct additional flows $\dot{\tilde{\vtheta}}_{\color{red}{t}}$, which depend on initial parameters, and showcase additional implicit regularisation effects.}
     \label{fig:idd_graphic_two_approaches}
\end{figure}

\textbf{Stochastic gradient descent (SGD)}. To model the implicit regularisation induced by the discretisation of the SGD update $\vtheta_t = \vtheta_{t -1} - h \nabla_{\vtheta} E(\vtheta_{t -1}; \vX_{t})$ corresponding to data batch $\vX_{t}$,
we can use the IGR flow induced at this time step:
\begin{align}
  \dot{\vtheta} = - \nabla_{\vtheta} E(\vtheta; \vX_{t}) - \frac h 4 \nabla_{\vtheta} \norm{\nabla_{\vtheta} E(\vtheta; \vX_{t})}^2.
\label{eq:igr_stochastic}
\end{align}
Since the vector field in Eq~\eqref{eq:igr_stochastic} is the negative gradient of the loss $E(\vtheta; \vX_{t}) +  \frac h 4 \norm{\nabla_{\vtheta} E(\vtheta; \vX_{t})}^2$, this reveals a local implicit regularisation which minimises the gradient norm $ \norm{\nabla_{\vtheta} E(\vtheta; \vX_{t})}^2$. It is not immediately clear, however, how to combine the IGR flows obtained for each SGD update in order to model the \textit{combined} effects of multiple SGD updates, each using a different batch. What are, if any, the implicit regularisation effects induced by two SGD steps, $\vtheta_t = \vtheta_{t -1} - h \nabla_{\vtheta} E(\vtheta_{t -1}; \vX_{t})$ and $\vtheta_{t+1} = \vtheta_{t} - h \nabla_{\vtheta} E(\vtheta_{t}; \vX_{t+1})$?~\citet{igr_sgd} find a modified flow in expectation over the shuffling of batches in an epoch, and use it to find implicit regularisation effects specific to SGD and study the effect of batch sizes in SGD. Since their approach works in expectation over an epoch, however, it does not account for the implicit regularisation effects of a smaller number of SGD steps, or account for the exact data batches used in the updates; we return to their results in the next section.
We take a different approach, and introduce a novel way to find implicit regularisers in SGD by revisiting the BEA proof structure and the  assumptions made thus far when using BEA.
Our approach can be summarised as follows:

\begin{remark}
Given the discrete update $\vtheta_t = \vtheta_{t-1} - h \nabla_{\vtheta}E (\vtheta_{t-1})$,
BEA constructs $\dot{\vtheta}$, such that $\norm{\vtheta(h; \vtheta_{t-1}) - \vtheta_t} \in \mathcal{O}(h^n)$ for a choice of $n \in \mathbb{N}, n \ge 3$. This translates into a constraint on the value of $\vtheta(h;\vtheta_{t-1})$.
Thus, BEA asserts only what the value of the correction terms $f_i$ in the vector field of the modified flow---see Eq~\eqref{eq:general_modified_vector_field}---is at $\vtheta_{t-1}$. 
Given the constraints on $f_i(\vtheta_{t-1})$, we can \textit{choose}  $f_i: \mathbb{R}^D \rightarrow \mathbb{R}^D$ in the vector field of the modified flow $\dot{\vtheta}$ to depend on the initial condition $\vtheta_{t-1}$. 
\label{remark:bea_proofs}
\end{remark}
As an example, in the proof by construction of the IGR flow (Section~\ref{sec:bea_proofs} in the SM), one obtains that if $\vtheta_t = \vtheta_{t-1} - h \nabla_{\vtheta}E (\vtheta_{t-1})$ and we want to find $\dot{\vtheta} = -\nabla_{\vtheta}E(\vtheta) + h f_1(\vtheta)$ such that $\norm{\vtheta(h; \vtheta_{t-1}) - \vtheta_t} \in \mathcal{O}(h^3)$, then $f_1({\color{red}{\vtheta_{t-1}}}) = - \frac{1}{2} \nabla_{\vtheta}^2 E({\color{red}{\vtheta_{t-1}}})\nabla_{\vtheta} E({\color{red}{\vtheta_{t-1}}})$ (Eq~\eqref{eq:igr_proof_value_iterate}). From there we (following~\citet{igr}) concluded that $f_1(\vtheta) = - \frac{1}{2} \nabla_{\vtheta}^2 E(\vtheta)\nabla_{\vtheta} E(\vtheta)$. But notice how BEA only \textit{sets a constraint on the value of the vector field at the initial point $\vtheta_{t-1}$}. If we allow the modified vector field to depend on the initial condition,  equally valid choices for $f_1$ are  $f_1(\vtheta) = - \frac{1}{2} \nabla_{\vtheta}^2 E(\vtheta)\nabla_{\vtheta} E({\color{red}{\vtheta_{t-1}}})$ or $f_1(\vtheta) =  -\frac{1}{2} \nabla_{\vtheta}^2 E({\color{red}{\vtheta_{t-1}}})\nabla_{\vtheta} E(\vtheta)$. \textit{By construction}, the above flows also have an error of $\mathcal{O}(h^3)$ after one GD step of learning rate $h$ with initial parameters $\vtheta_{t-1}$. The latter vector fields only describe the SGD update with initial parameters $\vtheta_{t-1}$ and thus they only apply to this \textit{specific SGD step}, though as previously noted that is also the case with the IGR flow due to the dependence on the data batch --- see Eq~\eqref{eq:igr_stochastic}. Their advantage lies in the ability to write modified losses when a modified vector field depending only on $\vtheta$ cannot be written as a gradient operator, as we shall see in the next sections. 
This observation leads us to the following remarks:
\begin{remark}
 There are multiple flows that lead to the same order in learning rate error after one discrete update. Many of these flows depend on the \textit{initial conditions} of the system, i.e. the initial parameters of the discrete update. We visualise this approach in Figure~\ref{fig:idd_graphic_two_approaches}.
\end{remark}

\begin{remark} Implicit in the choice of the existing BEA flows~\citep{igr,rosca2021discretisation,van2020simple} lies an assumption: that we are looking for the modified flows that hold at every training iteration. This is, however, challenged already in the case of SGD, where the modified flows depend on data batches, as shown in Eq~\eqref{eq:igr_stochastic}.
\end{remark}

\section{Implicit regularisation in multiple stochastic gradient descent steps}
\label{sec:modified_losses_sgd}

We now use the above observations to build modified flows that can be used to construct modified losses by writing the vector field of the flow as the negative gradient of a function; this enables us to capture the implicit regularisation effects of taking multiple SGD steps. 
We analyse $n$ SGD steps
\begin{align}
\vtheta_{t+\mu} = \vtheta_{t+\mu-1} - \nabla_{\vtheta} E(\vtheta_{t+\mu-1}; \vX^{t+\mu}), \hspace{3em} \mu \in \{0, \dots, n-1\},
\label{eq:sgd_multiple_updates_bea}
\end{align}
where $E(\vtheta_{t-1}; \vX) = \frac{1}{B}\sum_{i=1}^B E(\vtheta_{t-1}; \vx_i)$, with elements $\vx_i$ forming batch $\vX$.
We further denote $E(\vtheta; \{\vX^{t}, \dots, \vX^{t+n-1} \}) = \frac{1}{n} \sum_{\mu=0}^{n-1} E(\vtheta; \vX^{t+\mu})$, i.e. the average loss from the $n$ data batches.

The SGD updates in Eq~\eqref{eq:sgd_multiple_updates_bea} follow the gradient flow $\dot{\vtheta} = -\nabla_{\vtheta} E(\vtheta; \{\vX^{t}, \dots, \vX^{t+n-1} \})$ with an error of $\mathcal{O}(h^2)$;
 the effects of the mini-batches appear only at higher-order terms in learning rate $h$. We thus use BEA to find the modified flow that describes the SGD update with an error of  $\mathcal{O}(h^3)$,  with a vector field that can be written as a negative gradient, in order to find implicit regularisers (we provide proofs and the flows that construct the regularisers in the SM). This leads us to:

\begin{theorem} \label{thm:bea_sgd}
Denote $E(\vtheta; \{\vX^{t}, \dots, \vX^{t+n-1} \}) = \frac{1}{n} \sum_{\mu=0}^{n-1} E(\vtheta; \vX^{t+\mu})$, i.e. the average loss obtained from the $n$ data batches. Then the trajectory obtained by taking $n$ steps of SGD follows the trajectory of minimising the loss in continuous-time with $\mathcal{O}(h^3)$ error
\begin{align}
\tilde{E}(\vtheta) &= E(\vtheta; \{\vX^{t}, \dots, \vX^{t+n-1} \})
                  + \frac{n h}{4} \underbrace{\norm{ \nabla_{\vtheta}E(\vtheta; \{\vX^{t}, \dots, \vX^{t+n-1} \})}^2}_{\text{full batch norm regularisation}} \label{eq:sgd_norm_min}\\
                   & \quad - \frac{h}{n}\sum_{\mu =1 }^{n-1} \underbrace{\left[ \nabla_{\vtheta} E(\vtheta; \vX^{t+\mu})^T \left(\sum_{\tau = 0}^{\mu-1}    \nabla_{\vtheta}E({\color{red}{\vtheta_{t  -1}}}; \vX^{t+\tau})\right)\right]}_{\text{mini-batch gradient alignment}}.
\label{eq:modified_sup_learning}
\end{align}
\end{theorem}
We thus find the implicit regularisation effects induced by $n$ steps of SGD, capturing the \textit{importance of exact batches used and their order in Eq~\eqref{eq:modified_sup_learning}}. We note that without making use the observations regarding BEA in the previous section, and thus without the parameters $\vtheta_{t -1}$, a modified loss could not have been constructed outside of the full-batch case where one can recover the IGR flow.
The following remark immediately follows by setting $n=2$ in Eq~\eqref{eq:modified_sup_learning}:
\begin{remark} When taking a second SGD step, there is an implicit regularisation term maximising the dot product between the gradient at the current step and the gradient at the previous step: $\nabla_{\vtheta} E^T \nabla_{\vtheta} E(\vtheta_{t-1})$. This can be achieved by aligning the direction of the gradients between the two iterations or increasing the gradient norm. 
\label{rem:2sgd_steps}
\end{remark}

We compare this novel modified loss with the loss obtained by ignoring stochasticity and assuming $n$ full-batch updates; this entails using the IGR loss (proof for multiple steps in Section~\ref{sec:igr_multiple_steps})
\begin{align}
\tilde{E} = E(\vtheta;\{\vX^{t}, \dots, \vX^{t+n-1} \}) + \frac{h}{4} \|\nabla_{\vtheta} E(\vtheta; \{\vX^{t}, \dots, \vX^{t+n-1} \}) \|^2 .
\label{eq:igr_flow_mul}
\end{align}
The above modified losses show that both GD and SGD have a pressure to minimise the gradient norm  $\|\nabla_{\vtheta} E(\vtheta; \{\vX^{t}, \dots, \vX^{t+n-1} \})\|$. SGD leads to an additional regularisation effect capturing the importance of the order in which mini-batches are presented in training: maximising the dot product between gradients computed at the current parameters given a batch and the gradients computed at the initial parameters \textit{for all batches presented before the given batch}. While this can be achieved both by increasing the norm of the gradients or by aligning gradients with those at the initial iteration, we note that increasing the gradient norm is counter to the other regulariser induced by SGD, the gradient norm  minimisation effect shown in Eq~\eqref{eq:sgd_norm_min}.

Our approach is complementary to that of \citet{igr_sgd}, who obtain a relationship similar to Eq~\eqref{eq:modified_sup_learning} in expectation over all possible data batch shufflings $\sigma$ in an epoch---they describe expected value of the modified loss $\mathbb{E}_{\sigma} \left[\tilde{E}(\vtheta; \{\vX^{\sigma(t)}, \dots, \vX^{\sigma(t+n-1)} \}) \right]$:
\begin{align}
\mathbb{E_\sigma} \left[E_{sgd}(\vtheta) \right] &= E(\vtheta;\{\vX^{t}, \dots, \vX^{t+n-1} \}) + \frac{h}{4n} \sum_{k=0}^{n-1} \norm{\nabla_{\vtheta} E(\vtheta;\vX^{t+k})}^2. 
\end{align}
 Both approaches share the limitation that $nh$ needs to be suitably small for approximations to be relevant; we note that since we do not require $n$ to be the number of updates in an epoch---and we obtain interestingly regularisation effects for $n=2$, see Remark~\ref{rem:2sgd_steps}---this is less of an issue for our approach. Their approach has the advantage of finding an implicit regularisation effect that does not depend on the initial parameters. By depending on initial parameters, however, our approach does not require working in expectations and accounts for the exact batches used in the SGD updates. We hope this can be used to stabilise SGD over multiple steps, as has been done with a single GD step~\citep{rosca2022on}, that it can be used in continual and transfer learning~\citep{iman2022review,zhuang2020comprehensive,parisi2019continual}, as well as understanding the effects of the order of examples on model optimisation in online learning.

\section{Implicit regularisation in generally  differentiable two-player games}
\label{sec:dd_general_modified_losses}

\citet{rosca2021discretisation} expand the results of \citet{igr} to two-player games and use BEA to find distinct modified flows that  describe simultaneous and alternating GD. In zero-sum games, they construct modified losses from the continuous-time flows provided by BEA. These losses reveal sources of instability in zero-sum games, which in turn they use to construct stabilisation strategies that improve Generative Adversarial Network training with GD. Here, for simplicity we show results only for simultaneous updates when the players use the same learning rate $h$. Consider a game given by loss functions  $E_{\vphi}(\vphi, \vtheta): \mathbb{R}^m\times \mathbb{R}^n \rightarrow \mathbb{R}$ and $E_{\vtheta}(\vphi, \vtheta): \mathbb{R}^m\times \mathbb{R}^n \rightarrow \mathbb{R}$, respectively. 
\citet{rosca2021discretisation} show that simultaneous GD follows the modified continuous system
\begin{align}
 \dot{\vphi} &=  - \nabla_{\vphi}E_{\vphi}( \vphi, \vtheta)  + h \Big(\underbrace{- \frac{1}{4}  \nabla_{\vphi} \norm{ \nabla_{\vphi} E_{\vphi}}^2}_{\text{self term}} \underbrace{- \frac{1}{2} \jacparam{\vtheta}{\nabla_{\vphi} E_{\vphi}} \nabla_{\vtheta} E_{\vtheta}}_{\text{interaction term}}\Big), \label{eq:general_bae_formulation1}
 \\
 \dot{\vtheta}  &= - \nabla_{\vtheta}E_{\vtheta}( \vphi, \vtheta)   + h \Big(\underbrace{- \frac{1}{4}  \nabla_{\vtheta} \norm{ \nabla_{\vtheta} E_{\vtheta}}^2}_{\text{self term}}  \underbrace{- \frac{1}{2}
\jacparam{\vphi}{\nabla_{\vtheta} E_{\vtheta}} \nabla_{\vphi} E_{\vphi}}_{\text{interaction term}}\Big)\label{eq:general_bae_formulation2}
\end{align}
with a local error of order $\mathcal O(h^3)$, where we used their definition of self and interaction terms.
Using this framework, they construct \textit{modified loss functions} in the case of zero-sum ($E_{\vphi}( \vphi, \vtheta) = - E_{\vtheta}( \vphi, \vtheta)$) and common-payoff games ($E_{\vphi}( \vphi, \vtheta) = E_{\vtheta}( \vphi, \vtheta)$). Using their approach, however, since one cannot always write the interaction terms as gradient, \textit{one cannot always write the vector fields of the modified flows as a gradient for differentiable two-player games} and thus we cannot construct modified losses leading to implicit regularisers.
We now use the BEA approach we provide in this work to \textit{choose}  another set of modified flows with an error of $\mathcal{O}(h^3)$ to the discrete updates. We can now write modified losses for each GD iteration of a general two-player differentiable game, which depend on the iteration $t$) and describe the local trajectory of SGD up to $\mathcal{O}(h^3)$ (proof in SM):
\begin{align}
\tilde{E}_{\vphi, t} = E_{\vphi} +  h \Big(\underbrace{\frac{1}{2} \norm{\nabla_{\vphi}E_{\vphi}}^2}_{\text{self term}} + \underbrace{\frac{1}{2} \nabla_{\vtheta}  E_{\vphi}^T \nabla_{\vtheta} E_{\vtheta}({\color{red}\vphi_{t-1}, \vtheta_{t-1}}) }_{\text{interaction term}} \Big) \\
\tilde{E}_{\vtheta, t} = E_{\vtheta} +  h \Big(\underbrace{\frac{1}{2} \norm{\nabla_{\vtheta}E_{\vtheta}}^2}_{\text{self term}} + \underbrace{\frac{1}{2}  \nabla_{\vphi}  E_{\vtheta}^T \nabla_{\vphi} E_{\vphi}({\color{red}\vphi_{t-1}, \vtheta_{t-1}}}_{\text{interaction term}})\Big).
\end{align}

 The \textit{self terms} result in the implicit gradient regularisation found in supervised learning~\citep{igr} and  zero-sum games~\citep{rosca2021discretisation}: each player has an incentive to minimise its own gradient norm. The \textit{interaction term} for each player encourages minimising the dot product between the gradients of its loss with respect to the other player's parameters and the previous gradient update of the other player.
Consider the first player's interaction term, equal to $(-\nabla_{\vtheta} E_{\vphi}(\vphi_t, \cdot))^T (- \nabla_{\vtheta}E_{\vtheta}({\color{red}\vphi_{t-1}, \vtheta_{t-1}}))$, where $-E_{\vtheta}({\color{red}\vphi_{t-1}, \vtheta_{t-1}})$ is the previous update direction of $\vtheta$ aimed at minimising $E_{\vtheta}$. 
Its implicit regularisation effect depends on the functional form of $E_{\vphi}$ and $E_{\vtheta}$: if $\nabla_{\vtheta} E_{\vtheta}$ and $\nabla_{\vtheta} E_{\vphi}$ have aligned directions by construction, then the implicit regularisation effect nudges the first player's update towards a point in space where the second player's update changes direction; the opposite is true if $\nabla_{\vtheta} E_{\vtheta}$ and $\nabla_{\vtheta} E_{\vphi}$ are misaligned by construction.
 We work through this regularisation effect for a pair of commonly used GAN losses that do not form a zero-sum game (using the generator non-saturating loss \citep{goodfellow2014explaining}) in Appendix~\ref{app:non_sat_gan}, and contrast it with the zero-sum case (the saturating loss).

\section{Conclusion}
We provided a novel approach of interpreting backward error analysis and used it to find implicit regularisers induced by gradient descent in the single objective setting and in two-player games.
In the single objective case, we found implicit regularisation terms revealing importance of the alignment of gradients at the exact data batches used in multiple steps of stochastic gradient descent, while in two-player games we highlighted the need to examine the game structure in order to determine the effects of implicit regularisation.
We hope future work can empirically verify the effects of these implicit regularisers in deep learning.

\clearpage
\bibliography{main}
\clearpage
\appendix

\input{appendix}

\end{document}

%% file: macros.tex
\usepackage{amsmath, latexsym}
\usepackage{amsfonts}
\usepackage{mathtools}
\usepackage{dsfont}
\usepackage{booktabs}
\usepackage{xfrac}
\usepackage{bbm}

\usepackage[export]{adjustbox}

\usepackage{bm}
\usepackage{listings}

\usepackage[most]{tcolorbox}
\usepackage{xparse}
\usepackage{lipsum}
\usepackage{changepage}
\usepackage{enumitem}
\usepackage{pifont}

\usepackage{bm}
\usepackage{amssymb}
\usepackage{xifthen}

\usepackage{etoolbox} %

\definecolor{538blue}{RGB}{48,162,218}
\definecolor{538red}{RGB}{252,79,48}
\definecolor{538yellow}{RGB}{229,174,56}
\definecolor{538green}{RGB}{109,144,79}
\definecolor{538purple}{RGB}{129, 15, 124}
\definecolor{538grey}{RGB}{139,139,139}
\definecolor{538magenta}{RGB}{200,100,200}

\newcommand{\qedwhite}{\hfill \ensuremath{\Box}}

\newcommand\cut[1]{}

\usepackage[percent]{overpic}
\newcommand{\norm}[1]{\left\lVert#1\right\rVert}

\newcommand{\decorate}[2][]{%
    \ifthenelse{\isempty{#1}}{#2}{%
          \ifthenelse{\equal{\detokenize{#1}}{\detokenize{sn}}}{\hat{#2}}{\bar{#2}}%
          }}

\newcommand{\owntag}[1]{\stepcounter{equation}   \tag{#1, \theequation} }

\newcommand{\squishlist}{
   \begin{list}{$\bullet$}
    { \setlength{\itemsep}{0pt}      \setlength{\parsep}{3pt}
      \setlength{\topsep}{3pt}       \setlength{\partopsep}{0pt}
      \setlength{\leftmargin}{1.5em} \setlength{\labelwidth}{1em}
      \setlength{\labelsep}{0.5em} } }

\newcommand{\squishlisttwo}{
   \begin{list}{$\bullet$}
    { \setlength{\itemsep}{0pt}    \setlength{\parsep}{0pt}
      \setlength{\topsep}{0pt}     \setlength{\partopsep}{0pt}
      \setlength{\leftmargin}{2em} \setlength{\labelwidth}{1.5em}
      \setlength{\labelsep}{0.5em} } }

\newcommand{\squishend}{
    \end{list}  }

\newcommand{\myvec}[1]{\mathbf{#1}}

\newcommand{\myvecsym}[1]{\bm{#1}}

\newcommand{\vphi}{\myvecsym{\phi}}

\newcommand{\vtheta}{\myvecsym{\theta}}

\newcommand{\vx}{\myvec{x}}

\newcommand{\vz}{\myvec{z}}

\newcommand{\vX}{\myvec{X}}

\newcommand{\jacparam}[2]{\frac{d #2}{d #1}}

\newcommand{\evaljac}[3]{\jacparam{#1}{#2}\bigg\rvert_{#3}}

\newcommand{\evaljacdata}[4]{\evaljac{#1}{#2(\cdot; #4)}{#3}}

\newcommand{\jacthetaf}{\jacparam{\vtheta}{f}}
\newcommand{\jacphif}{\jacparam{\vphi}{f}}

\newcommand{\jacthetag}{\jacparam{\vtheta}{g}}
\newcommand{\jacphig}{\jacparam{\vphi}{g}}

\newcommand{\be}{\begin{equation}}
\newcommand{\ee}{\end{equation}}
\newcommand{\bea}{\begin{eqnarray}}
\newcommand{\eea}{\end{eqnarray}}
\newcommand{\beaa}{\begin{eqnarray*}}
\newcommand{\eeaa}{\end{eqnarray*}}

\DeclareMathAlphabet{\mathpzc}{OT1}{pzc}{m}{n}

%% file: appendix.tex
\appendix

\section{BEA proofs}
\label{sec:bea_proofs}
The general structure of BEA proofs is as follows: given a discrete update and a desired order of error $\mathcal{O}(h^n)$, then
\begin{itemize}
    \item Step 1. Expand the discrete updates in order to construct a relation between initial and final parameters up to order $\mathcal{O}(h^n)$.
    \item Step 2. Perform a Taylor expansion in $h$ of the modified flow in Eq~\eqref{eq:general_modified_vector_field}. Write each term in the Taylor expansion as a function of $\nabla_{\vtheta} E$ and the desired $f_i$ using the chain rule; group together terms of the same order in $h$ in the expansion.
    \item Step 3. Identify $f_i$ such that all terms of $\mathcal{O}(h^i)$ are equal to those in the discrete update obtained in Step 1 for $1 \le i \le n-2$.
\end{itemize}

We now exemplify how to use BEA to find the IGR flow in Eq~\eqref{eq:first_igr}~\citep{igr} describing one step of GD update $\vtheta_{t} = \vtheta_{t-1}- h \nabla_{\vtheta}E(\vtheta_{t-1})$ with an error of $\mathcal O(h^{3})$.  Since we are only looking for the first correction term, we only need to find $f_1$. For Step 1, there is nothing to do in this case, as the GD update satisfies the form we require. For Step 2, 
we perform a Taylor expansion to find the value of $\vtheta(h;\vtheta_{t-1})$ up to order $\mathcal O(h^{3})$.
We have
\begin{align}
\vtheta(h;\vtheta_{t-1}) = \vtheta_{t-1} + h \vtheta^{(1)}(\vtheta_{t-1}) + \frac{h^2}{2} \vtheta^{(2)}(\vtheta_{t-1}) +  \mathcal{O}(h^3).
\end{align}
We know by the definition of the modified vector field in Eq~\eqref{eq:general_modified_vector_field} that
\begin{align}
\vtheta^{(1)} = - \nabla_{\vtheta} E + h f_1({\vtheta}).
\end{align}
We can then use the chain rule to obtain
\begin{align}
\vtheta^{(2)} = \frac{d \left(- \nabla_{\vtheta} E + h f_1({\vtheta})\right)}{dt}  = - \frac{d \nabla_{\vtheta} E}{dt} + \mathcal{O}(h) = \nabla_{\vtheta}^2 E \nabla_{\vtheta} E + \mathcal{O}(h).
\end{align}
Thus,
\begin{align}
\vtheta(h;\vtheta_{t-1}) = \vtheta_{t-1} - h \nabla_{\vtheta} E(\vtheta_{t-1}) + h^2 f_1(\vtheta_{t-1}) + \frac{h^2}{2}  \nabla_{\vtheta}^2 E (\vtheta_{t-1})\nabla_{\vtheta} E (\vtheta_{t-1})+  \mathcal{O}(h^3).
\end{align}
We can then write
\begin{align}
\vtheta_t - \vtheta(h;\vtheta_{t-1}) &= \vtheta_{t-1} - h \nabla_{\vtheta} E(\vtheta_{t-1}) - \vtheta(h;\vtheta_{t-1}) \\
&= h^2 f_1(\vtheta_{t-1}) + \frac{h^2}{2}  \nabla_{\vtheta}^2 E (\vtheta_{t-1})\nabla_{\vtheta} E (\vtheta_{t-1})+  \mathcal{O}(h^3).
\end{align}
For the error to be of order $\mathcal{O}(h^3)$ the terms of order $\mathcal{O}(h^2)$ have to be $\mathbf{0}$, to match the GD update (Step 3).
This entails 
\begin{align}
f_1(\vtheta_{t-1}) =  -\frac{1}{2} \nabla_{\vtheta}^2 E(\vtheta_{t-1}) \nabla_{\vtheta} E(\vtheta_{t-1}),
\label{eq:igr_proof_value_iterate}
\end{align}
from which we conclude to $f_1 =  -\frac{1}{2} \nabla_{\vtheta}^2 E \nabla_{\vtheta} E $ leading to Eq~\eqref{eq:first_igr}.

\section{Supervised learning}

We denote by $\vtheta \in \mathbb{R}^D$ the parameters of the model and by $f: \mathbb{R}^D \rightarrow  \mathbb{R}^D$ the update function; we often consider $f = - \nabla_{\vtheta} E(\vtheta; \vx)$, where $E$ is the loss function but write $f$ for simplicity of notation. The Jacobian $\jacthetaf(\vtheta)$ is the $m\times m$ matrix
$\jacthetaf(\vtheta)_{i,j} = \left(\partial_{\theta_j}f_i\right)$ with $i = 1, \dots, D$ and  $j = 1, \dots, D$. Since we are interested in the stochastic setting, we write $f$ as an argument both of data and parameters: $f(\vtheta;\vx)$ denotes the application of $f$ at input $\vx$ using parameters $\vtheta$. We will denote as
\begin{align}
 E(\vtheta; \vX^{t}) = \frac 1 B \sum_{i=1}^{B}E(\vtheta; \vx_i^t) 
\end{align}
and 
\begin{align}
 f(\vtheta; \vX^{t}) &= - \frac 1 B \sum_{i=1}^{B} \nabla_{\vtheta} E(\vtheta; \vx_i^t) 
\label{eq:f_hat_def}\\
 f(\vtheta; \{\vX^{t}, ...\vX^{t+n-1}\}) &= \frac 1 n \sum_{i=0}^{n-1} f(\vtheta; \{\vX^{t+i}\}) \label{eq:multiple_batches_def}
\end{align}
as averages over batches for convenience and clarity.

We will use that 
\begin{align}
\jacparam{\vtheta}{\nabla_{\vtheta} E}{\nabla_{\vtheta} E} &=  \nabla_{\vtheta} \Big(\frac{\|\nabla_{\vtheta} E\|^2}2\Big) \label{eq:normssametheta}
\end{align}
repeatedly in our proofs.

Our goal is to find $\dot{\vtheta}$ such that the distance between $n$ steps of stochastic gradient descent with learning rate $h$ and $\dot{\vtheta}$ is of order $\mathcal{O}(h^3)$. 
We take the approach highlighted in Section~\ref{sec:bea_proofs}:
\begin{itemize}
  \item Step 1. We expand the $n$ discrete stochastic gradient descent updates up to $\mathcal{O}(h^3)$.
  \item Step 2. We expand change of the modified continuous updates of the form $\dot{\vtheta} = f + h f_1$ in time $nh$, where $f$ is the gradient function obtained using the concatenation of all batches used in the first step.
  \item  Step 3. We match the terms between the discrete and continuous updates of $\mathcal{O}(h^2)$  to find $f_1$.
\end{itemize}

\subsection{Two consecutive steps of stochastic gradient descent}

\textbf{Step 1: Expand the discrete updates.} \\
From the definition of stochastic gradient descent:
\begin{align}
 \vtheta_{t} &= \vtheta_{t-1} + h  f(\vtheta_{t-1}; \vX^{t}) \label{eq:sgd1} \\
 \vtheta_{t +1 } &= \vtheta_{t} + h   f(\vtheta_{t}; \vX^{t+1}) \label{eq:sgd2}
\end{align}
We expand the gradient descent steps and obtain:
\begingroup
\begin{align}
 \vtheta_{t +1 } &= \vtheta_{t} + h   f(\vtheta_{t}; \vX^{t+1}) \\
                &= \vtheta_{t-1} + h  f(\vtheta_{t-1}; \vX^{t}) + h   f(\vtheta_{t}; \vX^{t+1}) \owntag{From \eqref{eq:sgd1}} \\
                 &= \vtheta_{t-1} + h  f(\vtheta_{t-1}; \vX^{t}) + h   f(\vtheta_{t-1} + h  f(\vtheta_{t-1}; \vX^{t}); \vX^{t+1}) \owntag{From \eqref{eq:sgd2}} \\
                  &= \vtheta_{t-1} + h  f(\vtheta_{t-1}; \vX^{t}) + h   f(\vtheta_{t-1}; \vX^{t+1}) \\ &\hspace{2em}+  h^2 \evaljacdata{\vtheta}{f}{\vtheta_{t-1}}{\vX^{t+1}}  f(\vtheta_{t-1}; \vX^{t}) +  \mathcal{O}(h^3) \owntag{Taylor expansion}\\
                  &= \vtheta_{t-1} + 2h \left(\frac{1}{2} f(\vtheta_{t-1}; \vX^{t}) +  \frac{1}{2} f(\vtheta_{t-1}; \vX^{t+1})\right) \owntag{Grouping $f$ terms} \\ &\hspace{2em}+  h^2 \evaljacdata{\vtheta}{f}{\vtheta_{t-1}}{\vX^{t+1}}  f(\vtheta_{t-1}; \vX^{t}) +  \mathcal{O}(h^3)\\
                    &= \vtheta_{t-1} + 2h \!\!\!\underbrace{ f(\vtheta_{t-1}; \{\vX^{t}, \vX^{t+1} \})}_{\substack{\text{update with a batch obtained by}\\{\text{concatenating the two batches}}}} \!\! + \, h^2 \evaljacdata{\vtheta}{f}{\vtheta_{t-1}}{\vX^{t+1}}  f(\vtheta_{t-1}; \vX^{t})  +  \mathcal{O}(h^3)
\label{eq:two_steps_sgd}
\end{align}
\endgroup

\textbf{Step 2: Taylor expansion of modified flow} \\

We now expand what happens in a time of $2h$ in continuous-time, by using the form of $\dot{\vtheta}$ given by BEA
\begin{align}
\dot{\vtheta} =  f(\vtheta; \{\vX^{t}, \vX^{t+1} \}) + 2 h f_1(\vtheta; \{\vX^{t},\vX^{t+1}\} )
\end{align}
thus for any $\tau$
\begin{align}
{\vtheta}(\tau + 2h) &= {\vtheta}(\tau) + 2h \dot{\vtheta}(\tau) + 4h^2 \ddot{{\vtheta}}(\tau) + \mathcal{O}(h^3) \\
&= {\vtheta} + 2 h  f(\vtheta; \{\vX^{t}, \vX^{t+1} \}) \\ &\hspace{2em}+ 4 h^2 \left[f_1 + \frac{1}{2}  \jacparam{\vtheta}{f(\vtheta; \{\vX^{t}, \vX^{t+1} \})}  f(\vtheta; \{\vX^{t}, \vX^{t+1} \})\right] + \mathcal{O}(h^3) \label{eq:disp_sgd_flow}.
\end{align}

\textbf{Step 3: Matching terms of the second order} \\
From the above expansion of the discrete updates (Step 1) and of the continuous updates (Step 2) we can find the value of $f_1$ at the last initial parameters, $\vtheta_{t-1}$:
\begin{align}
4 h^2 \left[f_1 + \frac{1}{2}  \jacparam{\vtheta}{f(\vtheta; \{\vX^{t}, \vX^{t+1} \})}  f(\vtheta; \{\vX^{t}, \vX^{t+1} \})\right] = h^2 \evaljacdata{\vtheta}{f}{\vtheta_{t-1}}{\vX^{t+1}}  f(\vtheta_{t-1}; \vX^{t})
\end{align}
This leads to:
\begin{align}
f_1(\vtheta_{t-1}) &= \underbrace{- \frac{1}{2}  \evaljacdata{\vtheta}{f}{\vtheta_{t-1}}{\{\vX^{t}, \vX^{t+1} \}}  f(\vtheta_{t-1}; \{\vX^{t}, \vX^{t+1} \})}_{\text{the IGR full-batch drift term}} \owntag{From flow: \eqref{eq:disp_sgd_flow}} \\ &+ \frac 1 4 \evaljacdata{\vtheta}{f}{\vtheta_{t-1}}{\vX^{t+1}}  f(\vtheta_{t-1}; \vX^{t}) \owntag{From SGD: \eqref{eq:two_steps_sgd}}
\label{eq:sgd_f1_vheta_t}
\end{align}
From here we choose a function $f_1$ which satisfies the above constraint
\begin{align}
f_1(\vtheta) &= \underbrace{- \frac{1}{2}  \jacparam{\vtheta}{f(\vtheta; \{\vX^{t}, \vX^{t+1} \})}  f(\vtheta; \{\vX^{t}, \vX^{t+1} \})}_{\text{the IGR full-batch drift term}} \\ &\quad + \frac 1 4 \jacparam{\vtheta}{f(\vtheta; \vX^{t+1})}  f({\color{red}\vtheta_{t-1}}; \vX^{t}).
\end{align}
We have now found the modified flow $\dot{\vtheta}$, which by construction follows two Euler updates with an error of $\mathcal{O}(h^3)$. Note the use of the initial parameters (highlighted in red) in the flow's vector field; this choice is required to ensure we can write a modified loss function in the single objective optimisation setting by using $f = - \nabla_{\vtheta} E$:\textbf{}
\begin{align}
f_1(\vtheta) &=  - \frac{1}{2}  \jacparam{\vtheta}{\nabla_{\vtheta}E(\vtheta; \{\vX^{t}, \vX^{t+1} \})}  \nabla_{\vtheta}E(\vtheta; \{\vX^{t}, \vX^{t+1} \}) \\ &\quad+ \frac 1 4 \jacparam{\vtheta}{\nabla_{\vtheta}E(\vtheta; \vX^{t+1})}  \nabla_{\vtheta}E({\color{red}\vtheta_{t-1}}; \vX^{t}) \\
&= - \nabla_{\vtheta} \left(\frac{1}{4} \|\nabla_{\vtheta} E(\vtheta;\{\vX^{t}, \vX^{t+1} \}) \|^2  - \frac 1 4 \nabla_{\vtheta} E(\vtheta; \vX^{t+1})^T \nabla_{\vtheta} E({\color{red}{\vtheta_{t  -1}}}; \vX^t) \right) \owntag{using \eqref{eq:normssametheta}},
\end{align}
where $ E(\vtheta;\{\vX^{t}, \vX^{t+1} \}) = \frac {1} {2} \left( E(\vtheta;\vX^{t}) + E(\vtheta;\vX^{t+1})\right)$ is the loss $E$ evaluated at both mini-batches.
We are now ready to write the RHS of the modified flow as a gradient function
\begin{align}
\dot{\vtheta} = - \nabla_{\vtheta} \Bigg(&E(\vtheta;\{\vX^{t}, \vX^{t+1} \}) \\&+ 2h \left(\frac{1}{4} \|\nabla_{\vtheta} E(\vtheta;\{\vX^{t}, \vX^{t+1} \}) \|^2  - \frac 1 4 \nabla_{\vtheta} E(\vtheta; \vX^{t+1})^T \nabla_{\vtheta} E({\color{red}{\vtheta_{t  -1}}}; \vX^t) \right) \Bigg).
\end{align}
This leads to the modified loss
\begin{align}
\tilde{E} =& E(\vtheta;\{\vX^{t}, \vX^{t+1} \}) \\&+ 2h \left(\frac{1}{4} \|\nabla_{\vtheta} E(\vtheta;\{\vX^{t}, \vX^{t+1} \}) \|^2  - \frac 1 4 \nabla_{\vtheta} E(\vtheta; \vX^{t+1})^T \nabla_{\vtheta} E({\color{red}{\vtheta_{t  -1}}}; \vX^t) \right).
\end{align}

\subsection{Multiple steps of stochastic gradient descent}

We will now derive a similar result, for $n$ stochastic gradient descent steps, start at iteration $t$.

\noindent\textbf{Step 1: Expand discrete updates}
\begin{align}
 \vtheta_{t} &= \vtheta_{t-1} + h  f(\vtheta_{t-1}; \vX^{t}) \\
\dots \nonumber\\
 \vtheta_{t + n -1 } &= \vtheta_{t + n -2} + h   f(\vtheta_{t + n -2}; \vX^{t+n -1})
\end{align}
From Eq~\eqref{eq:two_steps_sgd}, we know that if we expand the first two steps we obtain
\begin{align}
 \vtheta_{t +1 } &= \vtheta_{t-1} + 2h  f(\vtheta_{t-1}; \{\vX^{t}, \vX^{t+1} \}) + h^2 \evaljacdata{\vtheta}{f}{\vtheta_{t-1}}{\vX^{t+1}}  f(\vtheta_{t-1}; \vX^{t})  +  \mathcal{O}(h^3) \label{eq:ih_multiple_steps}.
\end{align}
Then, by expanding the third step
\begingroup
\allowdisplaybreaks
\begin{align}
 \vtheta_{t +2 } &= \vtheta_{t +1} + h  f(\vtheta_{t+1}; \vX^{t+2}) \\
                 &=  \vtheta_{t-1} + 2h  f(\vtheta_{t-1}; \{\vX^{t}, \vX^{t+1} \}) + h^2 \evaljacdata{\vtheta}{f}{\vtheta_{t-1}}{\vX^{t+1}}  f(\vtheta_{t-1}; \vX^{t}) \owntag{Eq~\eqref{eq:ih_multiple_steps}} \\ &\hspace{1em}+    h  f(\vtheta_{t+1}; \vX^{t+2}) +  \mathcal{O}(h^3) \\
                 &=  \vtheta_{t-1} + 2h  f(\vtheta_{t-1}; \{\vX^{t}, \vX^{t+1} \}) + h^2 \evaljacdata{\vtheta}{f}{\vtheta_{t-1}}{\vX^{t+1}}  f(\vtheta_{t-1}; \vX^{t}) \\ &\hspace{1em}+   h  f\left({\vtheta_{t-1} + 2h  f(\vtheta_{t-1}; \{\vX^{t}, \vX^{t+1} \}) + h^2 \evaljacdata{\vtheta}{f}{\vtheta_{t-1}}{\vX^{t+1}}  f(\vtheta_{t-1}; \vX^{t})}; \vX^{t+2}\right) \nonumber \\&\hspace{1em} +  \mathcal{O}(h^3) \owntag{Eq~\eqref{eq:ih_multiple_steps}} \\
                 &=  \vtheta_{t-1} + 2h  f(\vtheta_{t-1}; \{\vX^{t}, \vX^{t+1} \}) + h^2 \evaljacdata{\vtheta}{f}{\vtheta_{t-1}}{\vX^{t+1}}  f(\vtheta_{t-1}; \vX^{t}) \\ 
                 &\hspace{1em}+   h  f(\vtheta_{t-1};\vX^{t+2})  \\
                 &\hspace{1em}+ h \evaljacdata{\vtheta}{f}{\vtheta_{t-1}}{\vX^{t+2}} \left(2h  f(\vtheta_{t-1}; \{\vX^{t}, \vX^{t+1} \}) + h^2 \evaljacdata{\vtheta}{f}{\vtheta_{t-1}}{\vX^{t+1}}  f(\vtheta_{t-1}; \vX^{t})\right) \nonumber \\&\hspace{1em} +  \mathcal{O}(h^3) \owntag{Taylor expansion}\\
                      &=  \vtheta_{t-1} + 2h  f(\vtheta_{t-1}; \{\vX^{t}, \vX^{t+1} \}) + h^2 \evaljacdata{\vtheta}{f}{\vtheta_{t-1}}{\vX^{t+1}}  f(\vtheta_{t-1}; \vX^{t}) \\ 
                 &\hspace{1em}+   h  f(\vtheta_{t-1};\vX^{t+2})  \\
                 &\hspace{1em}+ 2h^2  \evaljacdata{\vtheta}{f}{\vtheta_{t-1}}{\vX^{t+2}}  f(\vtheta_{t-1}; \{\vX^{t}, \vX^{t+1} \})  \owntag{Simplifying $h^3$ term} \\&\hspace{1em} +  \mathcal{O}(h^3) \\
                &=  \vtheta_{t-1} + 3h  f(\vtheta_{t-1}; \{\vX^{t}, \vX^{t+1}, \vX^{t+2} \}) \owntag{Grouping $f$ terms} \\ &\hspace{1em} + h^2 \evaljacdata{\vtheta}{f}{\vtheta_{t-1}}{\vX^{t+1}}  f(\vtheta_{t-1}; \vX^{t}) \\ 
                 &\hspace{1em}+ 2h^2  \evaljacdata{\vtheta}{f}{\vtheta_{t-1}}{\vX^{t+2}}  f(\vtheta_{t-1}; \{\vX^{t}, \vX^{t+1} \})  +  \mathcal{O}(h^3) \\
                  &=  \vtheta_{t-1} + 3h  f(\vtheta_{t-1}; \{\vX^{t}, \vX^{t+1}, \vX^{t+2} \}) + h^2 \evaljacdata{\vtheta}{f}{\vtheta_{t-1}}{\vX^{t+1}}  f(\vtheta_{t-1}; \vX^{t}) \\ 
                 &\hspace{1em}+ h^2  \evaljacdata{\vtheta}{f}{\vtheta_{t-1}}{\vX^{t+2}}  f(\vtheta_{t-1}; \vX^{t}) + h^2 \evaljacdata{\vtheta}{f}{\vtheta_{t-1}}{\vX^{t+2}}  f(\vtheta_{t-1}; \vX^{t+1}) \owntag{Eq~\eqref{eq:multiple_batches_def}} \\&\hspace{1em} +  \mathcal{O}(h^3) .
\end{align}
\endgroup
Thus by induction we get
\begin{align}
 \vtheta_{t +n-1 } &= \vtheta_{t-1} +  n h  f(\vtheta_{t-1}; \{\vX^{t}, \dots, \vX^{t+n-1} \})
                  \\ &+  h^2 \sum_{\tau = 0}^{n-1} \sum_{\mu =\tau+1 }^{n-1}  \evaljacdata{\vtheta}{f}{\vtheta_{t-1}}{\vX^{t+\mu}}  f(\vtheta_{t  -1}; \vX^{t+\tau}) + \mathcal{O}(h^3). \label{eq:sgd_n_exp}
\end{align}

\noindent\textbf{Step 2: Taylor expansion of modified flow} \\
\begin{align}
{\vtheta}(\tau + nh) &= {\vtheta} + n h  f(\vtheta; \{\vX^{t}, \dots, \vX^{t+n-1} \}) \\ &\quad+ n^2 h^2 \left[f_1 + \frac{1}{2} \jacparam{\vtheta}{f(\vtheta; \{\vX^{t}, \dots, \vX^{t+n-1} \})}  f(\vtheta; \{\vX^{t}, \dots, \vX^{t+n-1} \}) \right] \\& \quad+ \mathcal{O}(h^3)
\label{eq:taylor_exp_sgd}
\end{align}

\noindent\textbf{Step 3: Matching terms} \\
From the above expansion of the discrete updates (Step 1) and of the continuous updates (Step 2) we can find the value of $f_1$ at the last initial parameters, $\vtheta_{t-1}$ by matching the terms of order $h^2$:
\begin{align}
\sum_{\tau = 0}^{n-1} &\sum_{\mu =\tau+1 }^{n-1}  \evaljacdata{\vtheta}{f}{\vtheta_{t-1}}{\vX^{t+\mu}}  f(\vtheta_{t  -1}; \vX^{t+\tau}) \owntag{SGD: Eq~\eqref{eq:sgd_n_exp}} \\ &= n^2 \left[f_1 + \frac{1}{2} \jacparam{\vtheta}{f(\vtheta; \{\vX^{t}, \dots, \vX^{t+n-1} \})}  f(\vtheta; \{\vX^{t}, \dots, \vX^{t+n-1} \}) \right] \owntag{Flow: Eq~\eqref{eq:taylor_exp_sgd}}
\end{align}
Leading to
\begin{align}
f_1(\vtheta_{t-1}) &= \underbrace{- \frac{1}{2} \evaljacdata{\vtheta}{f}{\vtheta_{t-1}}{\{\vX^{t}, \dots, \vX^{t+n-1} \}}  f(\vtheta_{t-1}; \{\vX^{t}, \dots, \vX^{t+n-1} \})}_{\text{the usual drift term}} \\ &\quad + \frac{1}{n^2}\sum_{\tau = 0}^{n-1} \sum_{\mu =\tau+1 }^{n-1}  \evaljacdata{\vtheta}{f}{\vtheta_{t  -1}}{\vX^{t+\mu}}  f(\vtheta_{t  -1}; \vX^{t+\tau}) .
\label{eq:f_1_sgd_ours}
\end{align}
From here we choose a function $f_1$ which satisfies the above constraint.
\begin{align}
f_1(\vtheta) &= \underbrace{- \frac{1}{2} \jacparam{\vtheta}{f(\vtheta; \{\vX^{t}, \dots, \vX^{t+n-1} \})}  f(\vtheta; \{\vX^{t}, \dots, \vX^{t+n-1} \})}_{\text{the usual drift term}} \\ &\quad+ \frac{1}{n^2} \sum_{\tau = 0}^{n-1} \sum_{\mu =\tau+1 }^{n-1}  \jacparam{\vtheta}{f(\vtheta; \vX^{t+\mu})}  f({\color{red}{\vtheta_{t  -1}}}; \vX^{t+\tau}) 
\end{align}
We now replace $f = - \nabla_{\vtheta} E$, using that in this case $\jacthetaf = \nabla_{\vtheta}^2 E$,
\begin{align}
f_1(\vtheta_{t-1}) &= - \frac{1}{2} \jacparam{\vtheta}{\nabla_{\vtheta}E(\vtheta; \{\vX^{t}, \dots, \vX^{t+n-1} \})}  \nabla_{\vtheta}E(\vtheta; \{\vX^{t}, \dots, \vX^{t+n-1} \}) \\ &\quad + \frac{1}{n^2} \sum_{\tau = 0}^{n-1} \sum_{\mu =\tau+1 }^{n-1}  \jacparam{\vtheta}{\nabla_{\vtheta}E(\vtheta; \vX^{t+\mu})}  \nabla_{\vtheta}E({\color{red}{\vtheta_{t  -1}}}; \vX^{t+\tau}) \\
  &= - \nabla_{\vtheta}\Big(\frac{1}{4} \norm{ \nabla_{\vtheta}E(\vtheta; \{\vX^{t}, \dots, \vX^{t+n-1} \})}^2 \owntag{using \eqref{eq:normssametheta}} \\&\quad \quad \quad \quad - \frac{1}{n^2} \sum_{\tau = 0}^{n-1} \sum_{\mu =\tau+1 }^{n-1}  \nabla_{\vtheta} E(\vtheta; \vX^{t+\mu})^T \nabla_{\vtheta}E({\color{red}{\vtheta_{t  -1}}}; \vX^{t+\tau}) \Big)  .
\end{align}
This leads to the modified loss
\begin{align}
\tilde{E}(\vtheta) &= E(\vtheta; \{\vX^{t}, \dots, \vX^{t+n-1} \})\\ 
                   & \quad+  \frac{n h}{4} \norm{ \nabla_{\vtheta}E(\vtheta; \{\vX^{t}, \dots, \vX^{t+n-1} \})}^2 \\
                   & \quad- \frac{h}{n} \sum_{\tau = 0}^{n-1} \sum_{\mu =\tau+1 }^{n-1}  \nabla_{\vtheta} E(\vtheta; \vX^{t+\mu})^T \nabla_{\vtheta}E({\color{red}{\vtheta_{t  -1}}}; \vX^{t+\tau}).
\end{align}
Through algebraic manipulation we can write the above in the form
\begin{align}
\tilde{E}(\vtheta) &= E(\vtheta; \{\vX^{t}, \dots, \vX^{t+n-1} \})\\ 
                   & \quad + \frac{n h}{4} \norm{ \nabla_{\vtheta}E(\vtheta; \{\vX^{t}, \dots, \vX^{t+n-1} \})}^2 \\
                   & \quad - \frac{h}{n}\sum_{\mu =1 }^{n-1} \left[ \nabla_{\vtheta} E(\vtheta; \vX^{t+\mu})^T \left(\sum_{\tau = 0}^{\mu-1}    \nabla_{\vtheta}E({\color{red}{\vtheta_{t  -1}}}; \vX^{t+\tau})\right)\right].
\end{align}

We can now compare this with the modified loss we obtained by ignoring stochasticity and assuming both updates have been done with a full batch; this entails using the IGR loss:
\begin{align}
\tilde{E} = E(\vtheta;\{\vX^{t}, \dots, \vX^{t+n-1} \}) + \frac{h}{4} \|\nabla_{\vtheta} E(\vtheta; \{\vX^{t}, \dots, \vX^{t+n-1} \}) \|^2 
\end{align}

Thus, when using multiple batches, there is the additional pressure to maximise the dot product of the gradients obtained using the last $k$ batches that came before the current batch, evaluated at the parameters at which we started the $n$ iterations.

\subsection{Multiple steps of full-batch gradient descent}
\label{sec:igr_multiple_steps}

To contrast our results with multiple steps of gradient descent \textit{with the same batch}, we briefly show that the IGR flow follows gradient descent with error of order $\mathcal{O}(h^3)$ after $n$ gradient descent steps.

The proof steps follow the same approach as above, with two main differences: first, we assume all batches are the same, and second, we no longer fix the starting parameters when choosing $f_1$. With the same steps as before we obtain Eq~\eqref{eq:f_1_sgd_ours}:
\begin{align}
f_1(\vtheta_{t-1}) &= - \frac{1}{2} \evaljacdata{\vtheta}{f}{\vtheta_{t-1}}{\{\vX^{t}, \dots, \vX^{t+n-1} \}}  f(\vtheta_{t-1}; \{\vX^{t}, \dots, \vX^{t+n-1} \}) \\ &\quad + \frac{1}{n^2}\sum_{\tau = 0}^{n-1} \sum_{\mu =\tau+1 }^{n-1}  \evaljacdata{\vtheta}{f}{\vtheta_{t  -1}}{\vX^{t+\mu}}  f(\vtheta_{t  -1}; \vX^{t+\tau}) .
\end{align}
Assuming that all batches are equal, and using that $f$ is an empirical average, we obtain

\noindent ${f(\cdot; \{\vX^{t}, \dots, \vX^{t+n-1} \}) = f(\cdot; \vX^{t}) = f(\cdot; \vX^{t + i})}$, for any choice of $i$.
We then have 
\begin{align}
f_1(\vtheta) &= - \frac{1}{2} \jacparam{\vtheta}{f(\vtheta, \vX^{t})}  f(\vtheta; \vX^{t}) + \frac{1}{n^2}\sum_{\tau = 0}^{n-1} \sum_{\mu =\tau+1 }^{n-1}  \jacparam{\vtheta}{f(\vtheta, \vX^{t})}  f(\vtheta; \vX^{t}) = \\
   &= (- \frac{1}{2} + \frac{n(n-1)}{2 n^2})\jacparam{\vtheta}{f(\vtheta, \vX^{t})}  f(\vtheta; \vX^{t}) \\
   &= - \frac{1}{2n} \jacparam{\vtheta}{f(\vtheta, \vX^{t})}  f(\vtheta; \vX^{t})
\end{align}
Replacing $f = - \nabla_{\vtheta}E(\cdot; \vX^{t})$ into the form of the modified flow in Eq~\ref{eq:taylor_exp_sgd}, we obtain:
\begin{align}
\dot{\vtheta} &= - \nabla_{\vtheta}E(\vtheta; \vX^{t}) - \frac{nh}{2n} \nabla_{\vtheta}^2{E(\vtheta, \vX^{t})}  \nabla_{\vtheta}E(\vtheta; \vX^{t}) \\
&= - \nabla_{\vtheta}E(\vtheta; \vX^{t}) - \frac{h}{2} \nabla_{\vtheta}^2{E(\vtheta, \vX^{t})}  \nabla_{\vtheta}E(\vtheta; \vX^{t})
\end{align}
which is the standard IGR flow. Importantly, the flow  in the full-batch case \textit{does not depend on the number of iterations}.

\subsection{On learning rates and number of updates}
The important role of learning rates in BEA affects our results too: for our approximations to hold $nh$ has to be sufficiently small. If we adjust the learning rate by the number of updates, i.e. if we set the learning rate for stochastic gradient descent equal to $h/n$, where $h$ is the learning rate used by full-batch gradient descent, we obtain the same implicit regularisation coefficient  for the gradient norm $\|\nabla_{\vtheta} E(\vtheta;\{\vX^{t}, \dots, \vX^{t+n-1} \}) \|$ minimisation as the IGR flow in Eq~\ref{eq:igr_flow_mul}, and the main difference between the two modified losses is given by the dot product terms present in Eq~\eqref{eq:modified_sup_learning}. 

The number of updates, $n$, also plays an important role. While the dot product alignment term in Eq~\eqref{eq:modified_sup_learning} has a coefficient of $\frac{h}{n}$, there are $\frac{n(n-1)}{2}$ terms composing the term. Thus the magnitude of the dot product regularisation term can grow with $n$, but its effects strongly depend on the distribution of the gradients computed at different batches. For example, if gradients $\nabla_{\vtheta} E(\vtheta_{t-1};\vX^{t +i})$ are normally distributed with the mean at the full-batch gradient, i.e. $\nabla_{\vtheta} E(\vtheta_{t-1};\vX^{t +i}) \sim \mathcal{N}(\nabla_{\vtheta} E(\vtheta_{t-1}), \sigma^2)$, as the number of updates grows the regularisation effect in Eq~\ref{eq:modified_sup_learning} will result in a pressure to align mini-batch gradients with the full-batch gradient at the previous iteration parameters, $\nabla_{\vtheta} E(\vtheta_{t-1})$. Since our results hold for multiple values of $n$, empirical assessments need to be made to understand the interplay between the number of updates and the strength of dot product regularisation on training.

\subsection{Expectation over all shufflings}
\label{sec:sgd_igr_comp}
We contrast our results with those of~\citet{igr_sgd}, who construct a modified loss over an epoch of stochastic gradient descent in expectation over all possible shufflings of the batches in the epoch. We thus also take an expectation over all possible batch shufflings (but not the elements in the batch) in an epoch in Eq~\eqref{eq:modified_sup_learning}, and obtain
\begin{align}
\mathbb{E_\sigma} \left[E_{sgd}(\vtheta) \right]&= E(\vtheta; \{\vX^{t}, \dots, \vX^{t+n-1} \}) + \frac{nh}{4}  \norm{\nabla_{\vtheta} E(\vtheta;\{\vX^{t}, \dots, \vX^{t+n-1} \})}^2 
 \\& \hspace{1em}- \frac{h}{n} \mathbb{E_\sigma} \left[ \sum_{k=0}^{n-1} \nabla_{\vtheta} E(\vtheta;\vX^{t+k}) ^T\left(\sum_{i=0}^{k-1} \nabla_{\vtheta} E(\vtheta_{t-1}; \vX^{t+i})\right) \right] \\
                   &= E(\vtheta; \{\vX^{t}, \dots, \vX^{t+n-1} \}) + \frac{nh}{4}  \norm{\nabla_{\vtheta} E(\vtheta;\{\vX^{t}, \dots, \vX^{t+n-1} \})}^2 
                   \\& \hspace{1em}- \frac{h}{n} \frac{1}{2} \mathbb{E_\sigma} \left[\sum_{k=0}^{n-1} \sum_{i=0, i \ne k}^{n-1} \nabla_{\vtheta} E(\vtheta;\vX^{t+k}) ^T \nabla_{\vtheta} E(\vtheta_{t-1}; \vX^{t+i})\right] ,
\end{align}
where $\sigma$ denotes the set of all possible permutations of batches $\{1, ...n\}$. We used the symmetry of the permutation structure since for each permutation where $\sigma(i) < \sigma(j)$ there is also a permutation where $\sigma(i) > \sigma(j)$ by swapping the values of $\sigma(i)$ and $\sigma(j)$.
We expand the last term
\begin{align}
&\mathbb{E}_{\sigma} \left[E_{sgd} \right] 
                   =  E(\vtheta; \{\vX^{t}, \dots, \vX^{t+n-1} \}) + \frac{nh}{4}  \norm{\nabla_{\vtheta} E(\vtheta;\{\vX^{t}, \dots, \vX^{t+n-1} \})}^2 \\
                   &\hspace{1em} - \frac{h}{n} \frac{1}{2} \mathbb{E_\sigma} \left[\sum_{k=0}^{n-1} \sum_{i=0}^{n-1} \nabla_{\vtheta} E(\vtheta;\vX^{t+k}) ^T \nabla_{\vtheta} E(\vtheta_{t-1}; \vX^{t+i})\right]\\&\hspace{1em}  -  \frac{h}{n} \frac{1}{2} \mathbb{E_\sigma} \left[\sum_{k=0}^{n-1}  \nabla_{\vtheta} E(\vtheta;\vX^{t+k}) ^T \nabla_{\vtheta} E(\vtheta_{t-1}; \vX^{t+k})\right]
\end{align}
From here
\begin{align}
 \mathbb{E}_{\sigma} \left[E_{sgd} \right]                   &=  E(\vtheta; \{\vX^{t}, \dots, \vX^{t+n-1} \}) + \frac{nh}{4}  \norm{\nabla_{\vtheta} E(\vtheta;\{\vX^{t}, \dots, \vX^{t+n-1} \})}^2 
                    \\&\hspace{1em}- \frac{h}{2n} \nabla_{\vtheta} E(\vtheta; \{\vX^{t}, \dots, \vX^{t+n-1} \})^T \nabla_{\vtheta} E(\vtheta_{t-1}; \{\vX^{t}, \dots, \vX^{t+n-1} \} 
                   \\&\hspace{1em}  -  \frac{h}{2n}\left[\sum_{k=0}^{n-1}  \nabla_{\vtheta} E(\vtheta;\vX^{t+k}) ^T \nabla_{\vtheta} E(\vtheta_{t-1}; \vX^{t+k})\right].
\end{align}
We obtain that the pressure to minimise individual batch gradient norms is translated into a pressure to maximise the dot product between the gradients at the end of the epoch with those at the beginning of the epoch, both for each batch and for the entire dataset.

\section{Two-player games}
\label{app:non_sat_gan}
We denote by $\vphi \in \mathbb{R}^m$ and $\vtheta \in \mathbb{R}^n$ the parameters of the first and second player, respectively.
The players update functions will be denoted correspondingly by $f(\vphi, \vtheta): \mathbb{R}^m \times \mathbb{R}^n \rightarrow \mathbb{R}^m$ and by
$g(\vphi, \vtheta): \mathbb{R}^m \times \mathbb{R}^n \rightarrow \mathbb{R}^n$.
The Jacobian $\jacthetaf(\vphi, \vtheta)$ is the $m\times n$ matrix
$\jacthetaf(\vphi, \vtheta)_{i,j} = \left(\partial_{\theta_j}f_i\right)$ with $i = 1, \dots, m$ and  $j = 1, \dots, n$.

\citet{rosca2021discretisation} presented a framework for the quantification of numerical integration error in two player games.  They found distinct modified flows that  describe \textit{simultaneous} Euler updates and \textit{alternating} Euler updates.
That is, instead of using the original system
\begin{align}
 \dot{\vphi} &=  f( \vphi, \vtheta), \\
 \dot{\vtheta}  &= g( \vphi, \vtheta),
\end{align}\textbf{}
they find $f_1$, $g_1$, 
such that the modified continuous system
\begin{align}
 \dot{\vphi} &=  f( \vphi, \vtheta)  + h f_1( \vphi, \vtheta), \label{eq:general_bae_formulation1_app}\\
 \dot{\vtheta}  &= g( \vphi, \vtheta) + h g_1( \vphi, \vtheta) \label{eq:general_bae_formulation2_app}
\end{align}
follows the discrete steps of the method with a local error of order $\mathcal O(h^3)$.
More precisely, if $(\vphi_{t}, \vtheta_{t})$ denotes the discrete step of the method at time $t$ and $( \vphi(h),  \vtheta(h))$ corresponds to the continuous solution of the modified system above starting at $(\vphi_{t-1}, \vtheta_{t-1})$,  $\| \vphi_{t} -  \vphi( h) \| \textrm{ and }  \| \vtheta_{t} -  \vtheta( h) \|$
are of order $\mathcal O(h^3)$. We assume for simplicity that both players use the same learning rate $h$, but the same arguments can be made when they use different learning rates. If $f$, $g$ are negative gradient functions, i.e, $f = - \nabla_{\vphi} E_{\vphi}$, $g = - \nabla_{\vtheta} E_{\theta}$, the modified flows describe the behaviour of gradient descent in the game $\min_{\vphi} E_{\vphi}(\vphi, \vtheta)$ and $\min_{\vtheta} E_{\vtheta}(\vphi, \vtheta)$.

In their proof of Theorem 3.1, they have for simultaneous Euler updates 
\begin{align}
f_1(\vphi_{t-1}, \vtheta_{t-1}) &= - \frac{1}{2}  \jacphif(\vphi_{t-1}, \vtheta_{t-1}) f(\vphi_{t-1}, \vtheta_{t-1}) - \frac{1}{2}  \jacthetaf(\vphi_{t-1}, \vtheta_{t-1}) g(\vphi_{t-1}, \vtheta_{t-1}) \label{eq:proof_games_1}\\
g_1(\vphi_{t-1}, \vtheta_{t-1}) &= - \frac{1}{2} \jacphig(\vphi_{t-1}, \vtheta_{t-1}) f(\vphi_{t-1}, \vtheta_{t-1}) - \frac{1}{2}  \jacthetag(\vphi_{t-1}, \vtheta_{t-1}) g(\vphi_{t-1}, \vtheta_{t-1}) \label{eq:proof_games_2}.
\end{align}
From their, they choose $f_1$ and $g_1$ as:
\begin{align}
f_1(\vphi, \vtheta) &= - \frac{1}{2}  \jacphif(\vphi, \vtheta) f(\vphi, \vtheta) - \frac{1}{2}  \jacthetaf(\vphi, \vtheta) g(\vphi, \vtheta) \\
g_1(\vphi, \vtheta) &= - \frac{1}{2} \jacphig(\vphi, \vtheta) f(\vphi, \vtheta) - \frac{1}{2}  \jacthetag(\vphi, \vtheta) g(\vphi, \vtheta).
\end{align}

If we consider the situation of differentiable two-player games, we have that $f = -\nabla_{\vphi} E_{\vphi}$ and $g = -\nabla_{\vtheta} E_{\vtheta}$, where $E_{\vphi}(\vphi, \vtheta): \mathbb{R}^m\times \mathbb{R}^n \rightarrow \mathbb{R}$ and $E_{\vtheta}(\vphi, \vtheta): \mathbb{R}^m\times \mathbb{R}^n \rightarrow \mathbb{R}$ are the respective loss functions for the two players. Replacing this choice of $f$ and $g$ in the above, we obtain:
\begin{align}
 f_1 &= \underbrace{- \frac{1}{4}  \nabla_{\vphi} \norm{ \nabla_{\vphi} E_{\vphi}}^2}_{\text{self term}} \underbrace{- \frac{1}{2} \jacparam{\vtheta}{\nabla_{\vphi} E_{\vphi}} \nabla_{\vtheta} E_{\vtheta}}_{\text{interaction term}}, \\
 g_1 &= \underbrace{- \frac{1}{4}  \nabla_{\vtheta} \norm{ \nabla_{\vtheta} E_{\vtheta}}^2}_{\text{self term}}  \underbrace{- \frac{1}{2}
\jacparam{\vphi}{\nabla_{\vtheta} E_{\vtheta}} \nabla_{\vphi} E_{\vphi}}_{\text{interaction term}}
\end{align}

While the self terms can always be written as a gradient, whether or not the interaction term in the above formulation can only be written as a gradient for certain games --- \citet{rosca2021discretisation} focus on zero-sum and common-payoff games, where the interaction terms can be written as a gradient. Here, we use our novel interpretation of BEA and choose $f_1$ and $g_1$, now depending on the iteration number $t$, which will allow us to construct modified losses for general two-player games. Specifically, based on Eqs~\eqref{eq:proof_games_1} and~\eqref{eq:proof_games_2}, we choose
\begin{align}
f_{1, t}(\vphi, \vtheta) &= - \frac{1}{2}  \jacphif(\vphi, \vtheta) f(\vphi, \vtheta) - \frac{1}{2}  \jacthetaf(\vphi, \vtheta) g({\color{red}\vphi_{t-1}, \vtheta_{t-1}}) \\
g_{1, t}(\vphi, \vtheta) &= - \frac{1}{2}  \jacphig(\vphi, \vtheta) f({\color{red}\vphi_{t-1}, \vtheta_{t-1}}) - \frac{1}{2}  \jacthetag(\vphi, \vtheta) g(\vphi, \vtheta).
\end{align}
 Here, we treat $g(\vphi_{t-1}, \vtheta_{t-1})$ and $f(\vphi_{t-1}, \vtheta_{t-1})$ as constants; this allows us to write the interaction terms as negative gradients. We replace $f = - \nabla_{\vphi} E_{\vphi}$ and $g = - \nabla_{\vtheta} E_{\vtheta}$  in $f_{1, t}$ and $g_{1, t}$ and write the drift terms as gradient functions:
\begin{align}
f_{1, t}(\vphi, \vtheta) &= - \frac{1}{2} \jacparam{\vphi}{\nabla_{\vphi} E_{\vphi}} \nabla_{\vphi} E_{\vphi}  - \frac{1}{2} \jacparam{\vtheta}{\nabla_{\vphi} E_{\vphi}} \nabla_{\vtheta} E_{\vtheta}(\vphi_{t-1}, \vtheta_{t-1}) \\
&= - \frac{1}{2} \nabla_{\vphi} \norm{\nabla_{\vphi}E_{\vphi}}^2 - \frac{1}{2} \nabla_{\vphi}\left(\nabla_{\vtheta}  E_{\vphi} ^T \nabla_{\vtheta} E_{\vtheta}(\vphi_{t-1}, \vtheta_{t-1}) \right) \\
&= - \nabla_{\vphi} \left(\frac{1}{2} \norm{\nabla_{\vphi}E_{\vphi}}^2 + \frac{1}{2} \nabla_{\vtheta}  E_{\vphi}^T  \nabla_{\vtheta} E_{\vtheta}(\vphi_{t-1}, \vtheta_{t-1}) \right) \\
g_{1, t}(\vphi, \vtheta) &= - \nabla_{\vtheta} \left(\frac{1}{2} \norm{\nabla_{\vtheta}E_{\vtheta}}^2 + \frac{1}{2} \nabla_{\vphi}  E_{\vtheta}^T \nabla_{\vphi} E_{\vphi}(\vphi_{t-1}, \vtheta_{t-1}) \right).
\end{align}
Replacing the above in the modified flows given by BEA (Eqs~\eqref{eq:general_bae_formulation1_app} and~\eqref{eq:general_bae_formulation2_app})
\begin{align}
 \dot{\vphi} &=  f( \vphi, \vtheta)  + h f_1( \vphi, \vtheta) \\
              &= - \nabla_{\vphi} \left(E_{\vphi} +  h \left(\frac{1}{2} \norm{\nabla_{\vphi}E_{\vphi}}^2 + \frac{1}{2} \nabla_{\vtheta}  E_{\vphi}^T \nabla_{\vtheta} E_{\vtheta}(\vphi_{t-1}, \vtheta_{t-1}) \right)   \right)  \label{eq:disc_int_app}\\
 \dot{\vtheta}  &= g( \vphi, \vtheta) + h g_1( \vphi, \vtheta) \\
                &= - \nabla_{\vtheta} \left(E_{\vtheta} +  h \left(\frac{1}{2} \norm{\nabla_{\vtheta}E_{\vtheta}}^2 + \frac{1}{2} \nabla_{\vphi}  E_{\vtheta} ^T \nabla_{\vphi} E_{\vphi}(\vphi_{t-1}, \vtheta_{t-1}) \right)\right).
\end{align}

\subsection{Non-saturating and saturating GAN losses}

We now investigate the effect of implicit regularisation found in Section~\ref{sec:dd_general_modified_losses} on GANs, specifically, the effect of the interaction terms, which takes the form of a dot product. We will denote the first player, the discriminator, as $D$, parametetrised by $\vphi$, and the generator as $G$, parametrised by $\vtheta$. We denote the data distribution as $p^*(\vx)$ and the latent distribution $p(\vz)$.
Consider the non-saturating GAN loss described by \citet{goodfellow2014generative}:
\begin{align}
    E_{\vphi}(\vphi, \vtheta) &= \mathbb{E}_{p^*(\vx)} \log D(\vx; \vphi) + \mathbb{E}_{p(\vz)} \log (1 - D(G(\vz; \vtheta); \vphi)) \\
     E_{\vtheta}(\vphi, \vtheta) &=  \mathbb{E}_{p(\vz)} - \log D(G(\vz; \vtheta); \vphi) 
\end{align}

Consider the interaction term for the discriminator (see Eq \eqref{eq:disc_int_app}), and replace the definitions of the above loss functions:
\begin{align}
\nabla_{\vtheta} & E_{\vphi}^T \nabla_{\vtheta} E_{\vtheta}(\vphi_{t-1}, \vtheta_{t-1}) \\ &= \nabla_{\vtheta} \left(\mathbb{E}_{p^*(\vx)} \log D(\vx; \vphi) + \mathbb{E}_{p(\vz)} \log (1 - D(G(\vz; \vtheta); \vphi)\right)^T \nabla_{\vtheta} E_{\vtheta}(\vphi_{t-1}, \vtheta_{t-1}) \\
 &= \nabla_{\vtheta} \left(\mathbb{E}_{p(\vz)} \log (1 - D(G(\vz; \vtheta); \vphi)\right)^T \nabla_{\vtheta} E_{\vtheta}(\vphi_{t-1}, \vtheta_{t-1}) \\
  &=\left(\mathbb{E}_{p(\vz)}  \nabla_{\vtheta}  \log (1 - D(G(\vz; \vtheta); \vphi)\right)^T \nabla_{\vtheta} E_{\vtheta}(\vphi_{t-1}, \vtheta_{t-1}) \\
    &=\left(- \mathbb{E}_{p(\vz)} \frac{1}{1 - D(G(\vz; \vtheta); \vphi)} \nabla_{\vtheta}  D(G(\vz; \vtheta); \vphi)\right)^T \nabla_{\vtheta} E_{\vtheta}(\vphi_{t-1}, \vtheta_{t-1}) \\
    &=\left(- \mathbb{E}_{p(\vz)} \frac{1}{1 - D(G(\vz; \vtheta); \vphi)} \nabla_{\vtheta}  D(G(\vz; \vtheta); \vphi)\right)^T  \\ & \hspace{5em} \left( - \mathbb{E}_{p(\vz)} \frac{1}{D(G(\vz; \vtheta_{t-1}); \vphi_{t-1})} \nabla_{\vtheta}  D(G(\vz; \vtheta_{t-1}); \vphi_{t-1})\right) \\
    &=\left(\mathbb{E}_{p(\vz)} \frac{1}{1 - D(G(\vz; \vtheta); \vphi)} \nabla_{\vtheta}  D(G(\vz; \vtheta); \vphi)\right)^T \\ & \hspace{5em} \left( \mathbb{E}_{p(\vz)} \frac{1}{D(G(\vz; \vtheta_{t-1}); \vphi_{t-1})} \nabla_{\vtheta}  D(G(\vz; \vtheta_{t-1}); \vphi_{t-1})\right)
\end{align}
where the expectations can be evaluated at the respective mini-batches used in the updates for iterations $t$ and $t-1$, respectively. Consider $\vz^i_t$ the latent variable with index $i$ in the batch at time $t$. Then the above is approximated as
\begin{align}
 \frac{1}{B^2} \sum_{i,j=1}^B &c^{non-sat}_{i,j}  \nabla_{\vtheta}  D(G(\vz^i_t; \vtheta); \vphi)^T   \nabla_{\vtheta}  D(G(\vz^j_{t-1}; \vtheta_{t-1}); \vphi_{t-1}), \hspace{3em} \text{with} \\
    &c^{non-sat}_{i,j} = \frac{1}{1 - D(G(\vz^i_t; \vtheta); \vphi)} \frac{1}{D(G(\vz^j_{t-1}; \vtheta_{t-1}); \vphi_{t-1})} 
\end{align}
Thus, the strength of the regularisation---$c^{non-sat}_{i,j}$---depends on how \textit{confident} the discriminator is. In particular, this implicit regularisation encourages the discriminator update into a new set of parameters where the gradient $\nabla_{\vtheta}  D(G(\vz^i_t; \vtheta); \vphi)$ \textit{points away} from  ${\nabla_{\vtheta}  D(G(\vz^j_{t-1}; \vtheta_{t-1}); \vphi_{t-1})}$ when $c_{i,j}$ is large. This occurs for $\vz^i_t$ where the discriminator is fooled by the generator---i.e. $1 - D(G(\vz_i; \vtheta); \vphi)$ is close to 0 and  $\frac{1}{1 - D(G(\vz^i_t; \vtheta); \vphi)}$ is large---and samples $\vz^j_{t-1}$ where the discriminator was correct---$D(G(\vz^j_{t-1}; \vtheta_{t-1}); \vphi_{t-1})$ low and thus $\frac{1}{D(G(\vz^j_{t-1}; \vtheta_{t-1}); \vphi_{t-1})}$ is large---at the previous iteration. This can be seen as beneficial regularisation for the generator, as it can be helpful at the next generator update by ensuring the update direction $- {\nabla_{\vtheta} E_{\vtheta} = \frac{1}{B}\sum_{i=1}^{B} \frac{1}{D(G(\vz_t^i; \vtheta); \vphi)} \nabla_{\vtheta}  D(G(\vz_t^i; \vtheta); \vphi)}$ is adjusted accordingly to the discriminator's output. We note, however, that regularisation might not have a strong  effect, as gradients $\nabla_{\vtheta}  D(G(\vz^i_t; \vtheta); \vphi)$ that have a high weight in the generator's update are those where the discriminator is correct---i.e. $\frac{1}{D(G(\vz^i_t; \vtheta); \vphi)}$ is large--- but there the factor $\frac{1}{1 - D(G(\vz^i_t; \vtheta); \vphi)}$ in $c^{non-sat}_{i,j}$ will be low. This is inline with the empirical results of~\citet{rosca2021discretisation}, who show that  for the non-saturating loss discretisation error does not have a strong effect on performance.

We can contrast this with what regularisation we obtain when using the saturating loss \citep{goodfellow2014generative}, where $E_{\vtheta} = - \mathbb{E}_{p(\vz)} \log (1 - D(G(\vz; \vtheta); \vphi))$, where we obtain the following implicit regulariser
\begin{align}
   \frac{1}{B^2}  \sum_{i,j=1}^B &c^{sat}_{i,j} \nabla_{\vtheta}  D(G(\vz^i_t; \vtheta); \vphi)^T   \nabla_{\vtheta}  D(G(\vz^j_{t-1}; \vtheta_{t-1}); \vphi_{t-1}), , \hspace{3em} \text{with}\\
    &c^{sat}_{i,j} = \frac{1}{1 - D(G(\vz^i_t; \vtheta); \vphi)} \frac{1}{1-D(G(\vz^j_{t-1}; \vtheta_{t-1}); \vphi_{t-1})} 
\end{align}
Here, $c^{sat}_{i,j}$ is high for $\vz^i_t$ and $\vz^j_{t-1}$ where the generator was fooling the discriminator---i.e. low $1-D(G(\vz^j_{t-1}; \vtheta_{t-1}); \vphi_{t-1})$ and $1 - D(G(\vz^i_t; \vtheta); \vphi)$. Thus, instead of moving $\nabla_{\vtheta}  D(G(\vz^i_t; \vtheta); \vphi)$ away from directions where the generator was doing poorly previously as is the case for the non-saturating loss, \textit{it is moving it away from directions where the generator was performing well}, which could lead to instabilities or loss of performance. Moreover, unlike for the non-saturating loss, the implicit regularisation will have a strong effect on the update $- {\nabla_{\vtheta} E_{\vtheta} = \mathbb{E}_{p(\vz)} \frac{1}{1-D(G(\vz; \vtheta); \vphi)} \nabla_{\vtheta}  D(G(\vz; \vtheta); \vphi)}$, since gradients $\nabla_{\vtheta}  D(G(\vz; \vtheta); \vphi)$ that have a high weight in the generator's update are those where $c_{i,j}$ is high. This is inline with \citet{rosca2021discretisation}, who show that in zero-sum games (such as the one induced by the saturating loss), the regularisation induced by the discretisation error of gradient descent is strong, and can hurt performance and stability.